\documentclass[10pt,twocolumn,letterpaper]{article}

\usepackage{cvpr}              
\usepackage[dvipsnames]{xcolor}

\usepackage{times}
\usepackage{soul}
\usepackage[utf8]{inputenc}
\usepackage{graphicx}
\usepackage{amsmath}
\usepackage{amsthm}
\usepackage{booktabs}
\usepackage{algorithm}
\usepackage{algorithmic}
\usepackage[switch]{lineno}

\usepackage{multirow}
\usepackage{adjustbox}
\usepackage{amssymb,amsfonts}

\usepackage{listings}
\usepackage{kotex}
\usepackage[accsupp]{axessibility}

\newcommand{\Real}{\mathbb{R}}

\newcommand{\StateSet}{\mathcal{S}}
\newcommand{\ActionSet}{\mathcal{A}}
\newcommand{\Dynamics}{\mathcal{P}}
\newcommand{\RewardFunc}{R}
\newcommand{\oursol}{\textsf{MoDeC}}
\newcommand{\textoursol}{MoDeC}

\DeclareMathOperator*{\argmax}{argmax}

\DeclareMathOperator*{\expectation}{\mathbb{E}}

\newcommand{\BaseNet}{\pi_\text{base}}
\newcommand{\PTBaseNet}{\bar{\pi}_\text{base}}
\newcommand{\IterMSNet}{\pi_\text{ims}}
\newcommand{\MSNet}{\pi_\text{ms}}
\newcommand{\ModeLearner}{\pi_\text{da}}

\definecolor{cvprblue}{rgb}{0.21,0.49,0.74}
\usepackage[pagebackref,breaklinks,colorlinks,citecolor=cvprblue]{hyperref}

\def\paperID{17148} 
\def\confName{CVPR}
\def\confYear{2024}

\title{Model Adaptation for Time Constrained Embodied Control}
\author{Jaehyun Song\thanks{Equal contribution.}, 
\setcounter{footnote}{0}
Minjong Yoo\footnotemark{},
\setcounter{footnote}{1}
Honguk Woo\thanks{Honguk Woo is the corresponding author.}\\
Department of Computer Science and Engineering, Sungkyunkwan University\\
{\tt\small \{s7159540, mjyoo2, hwoo\}@skku.edu }
}

\begin{document}
\maketitle

\begin{abstract}
When adopting a deep learning model for embodied agents, it is required that the model structure be optimized for specific tasks and operational conditions. Such optimization can be static such as model compression or dynamic such as adaptive inference. 
Yet, these techniques have not been fully investigated for embodied control systems subject to time constraints, which necessitate sequential decision-making for multiple tasks, each with distinct inference latency limitations. 
In this paper, we present $\oursol$, a time constraint-aware embodied control framework using the modular model adaptation.
We formulate model adaptation to varying operational conditions on resource and time restrictions as dynamic routing on a modular network, incorporating these conditions as part of multi-task objectives. 
Our evaluation across several vision-based embodied environments demonstrates the robustness of $\oursol$, showing that it outperforms other model adaptation methods in both performance and adherence to time constraints in robotic manipulation and autonomous driving applications.  
\end{abstract}

\section{Introduction}
In the literature on embodied artificial intelligence (embodied AI), where deep learning models have been increasingly adopted, optimizing the deep learning model structure for specific tasks and operational conditions becomes crucial. 
Several studies focus on static model optimization and compression~\cite{wang2022learnable, liang2023less, haldar2023polytask}. 
On the other hand, there are a few studies that investigate dynamic model structures, which are designed to adapt more effectively to conditions that change over time~\cite{liu2018dynamic, yang2020multi, cai2021dynamic, li2021dynamic, bolukbasi2017adaptive}. These dynamic approaches, often known as adaptive inference, are particularly beneficial for embodied environments where surrounding conditions evolve and the agent model's capabilities to adapt in real-time are essential.
Yet, a substantial gap remains in the exploration of these approaches within the context of embodied control systems, especially those operating under strict time constraints and involving more than a single task.   
Specifically, the systems are distinct in their requirement for sequential decision-making across multiple tasks, each with its own latency limitations. For example, consider an autonomous driving agent. This agent must continuously make driving control decisions in response to moving obstacles. Each decision must be made rapidly and accurately, as any delay can significantly impact driving safety and efficiency due to the time-sensitive nature of driving tasks.  

To address the challenges faced in such time-constrained embodied environments, we present $\oursol$ (MoDel adaptation for time constrained Embodied Control), a novel modular network framework with efficient and adaptive inference capabilities. 
In embodied AI, where models are deployed on specific hardware with different limitations, the ability to rapidly adapt to both operational resource conditions and time constraints is crucial. Recognizing this, we employ a constraint-aware modular model architecture, which can transform the procedure of time-sensitive inference into effective module selection within a single modular network that can be deployed on different target devices. 
This approach enables dynamic model adaptation to varied operational conditions in a sample-efficient way.
We also use a meta-learning scheme combined with knowledge distillation for restructuring the module selection in non-iterative time-sensitive computations.  

Through intensive experiments with several embodied control scenarios such as robot manipulation tasks in Meta-world~\cite{yu2020meta}, autonomous driving tasks in CARLA~\cite{dosovitskiy2017carla}, and object navigation tasks in AI2THOR~\cite{kolve2017ai2}, we demonstrate that our $\oursol$ framework is applicable for different time constraints and devices, achieving robust adaptation performance in terms of both constraint satisfaction and model accuracy.  
For instance, $\oursol$ shows a performance gain of 14.4\%  in success rates over the most competitive baseline, DS-Net~\cite{li2021dynamic} for autonomous driving tasks in CARLA, while it keeps the violation of time constraints to be less than 1\%. 
The main contributions of this paper are summarized as follows.
\begin{itemize}
    \item  We present the constraint-aware modular model framework $\oursol$, specifically designed for dynamic multi-task model adaptation to time constraints and device resource specifications. 
    \item We devise an efficient joint learning algorithm for optimizing the combinatorial module selection in a model and stabilizing the model against the large action space and non-stationarity problems. 
    We also employ the distillation-based inference optimization. 
    \item We evaluate the framework with several embodied environments and embedded devices, demonstrating its robustness and adaptability in terms of time-sensitive inference performance upon a wide range of tasks and operational conditions.
\end{itemize}

\section{Related Work}

\noindent \textbf{Embodied AI.}
Achieving an embodied agent requires learning complex and diverse tasks, as well as adapting to a constantly changing, real-world environment. Many researchers focused on complex tasks in embodied environments, including object navigation~\cite{wortsman2019learning, campari2020exploiting, chaplot2020object, du2020learning, wahid2021learning}, and embodied question and answering~\cite{das2018embodied, yu2019multi, tan2023knowledge, luo2023robust}, as well as model transfer from simulation to deployment environments~\cite{gordon2019splitnet, li2020unsupervised, kotar2022interactron}.
Specifically, \cite{gordon2019splitnet} introduced SplitNet, decoupling visual perception and policy learning. By decomposing the network architecture into a visual encoder and a task decoder, it allows for rapid adaptation to new domains and vision tasks.
\citet{li2020unsupervised} also presented a learning framework that can adjust a learned policy to the target environment that differs from the training environment, utilizing unlabeled data from the target. 
While sharing the similar goal to adapt to different embodied environment conditions with the prior works, we focus on dynamic model adaptation to both time constraints and device limitations in the context of multi-task policy learning and inference.  

\noindent \textbf{Real-time model inference.}
Several works have been introduced in the realm of real-time model inference~\cite{teerapittayanon2016branchynet, bolukbasi2017adaptive, liu2018dynamic, cai2021dynamic, hua2019channel, xia2021fully, li2021dynamic}.
Specifically,~\citet{cai2021dynamic} explored the trade-off between model performance and inference efficiency, by selecting certain nodes of the network and some of various filter-size CNN layers. 
\citet{li2021dynamic} introduced DS-Net, where weight ratios within each convolution neural network determine the network slicing for optimized inference. 
Unlike these works that enable the model adaptation for a given static condition, our work considers instance-wise operational conditions that can be given as input to the model for adaptation.    

\noindent \textbf{Model adaptation.}
Various works for quickly adapting a model to different environment features have been presented; e.g., for task difficulty levels~\cite{huang2017multi, wang2020dual, li2021dynamic, chang2021reducing, raychaudhuri2022controllable}, unseen tasks~\cite{fried2018speaker, yu2020take, gao2022dialfred, liu2021vision, li2022envedit}, and embodied properties~\cite{nair2022r3m, majumdar2023we}. 
\citet{wang2020dual} proposed a framework that selectively skips CNN layers and channels within layers based on the task difficulty. They also exploited an early exit mechanism for resource-constrained inference.  
\citet{han2022latency} introduced a training algorithm and architecture that can adjust the latency by pixel-wise masking of CNNs, employing the latency prediction model and dynamic adjustment of hyperparameters for different devices.
In line with these dynamic model adaptation works, we employ a modular network architecture and learning algorithm specifically designed for embodied control systems with multiple tasks, time constraints, and different device specifications.

\begin{figure}[t] 
    \centering
        \includegraphics[width=0.93\columnwidth]{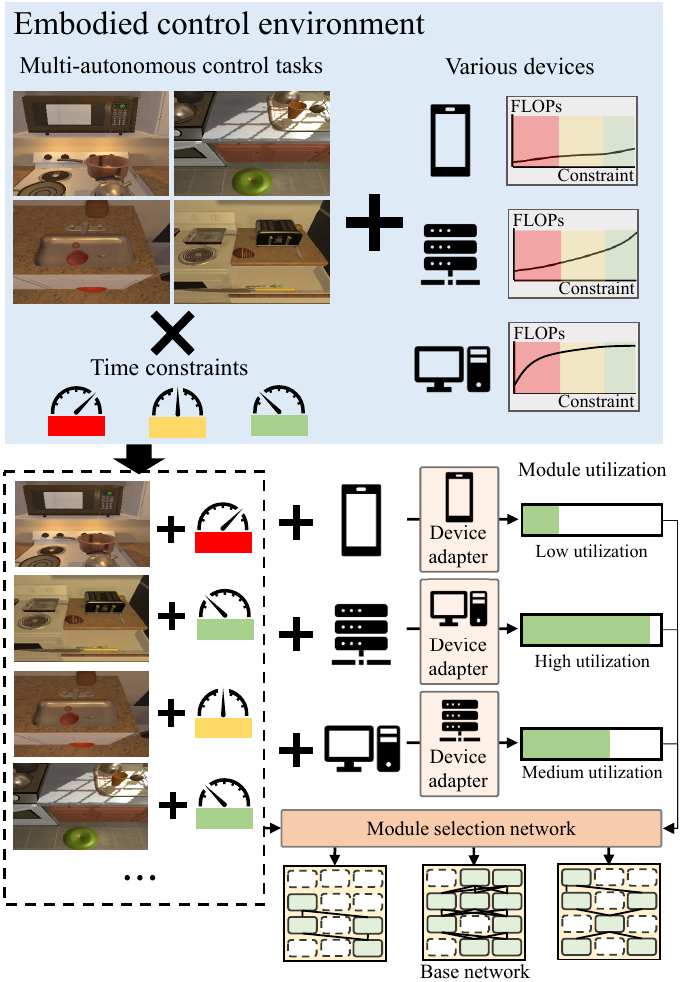}
    \caption{Overall Architecture} 
    \label{fig:Fig1}
    \vspace{-0.1in}
\end{figure}

\section{Approach}
\subsection{Problem Formulation}
We consider reinforcement learning (RL) for multi-autonomous control tasks in embodied environments. As an individual task is formulated as a single Markov decision process (MDP) $\mathcal{M}$, a multi-task MDP is equal to a family of MDPs $\{\mathcal{M}_i = (\StateSet, \ActionSet, \Dynamics_i, \RewardFunc_i, \gamma)\}_i$.  
$\StateSet$ is a set of states, $\ActionSet$ is a set of actions, $\Dynamics:\StateSet \times \ActionSet \times \StateSet \rightarrow [0, 1]$ is a transition probability, $\RewardFunc: \StateSet \times \ActionSet \rightarrow \Real$ is a reward function, and $\gamma$ is a discount factor. 
In multi-task RL, task information is used to reformulate a family of MDPs in a single MDP. Accordingly, given a task index $i \in I$, such a reformulated MDP can be represented as $(\StateSet \times I, \ActionSet, \Dynamics_I, \RewardFunc_I, \gamma)$ where $\Dynamics_I((s, i), a) = \Dynamics_i(s, a)$ and $\RewardFunc_I((s, i), a) = \RewardFunc_i(s, a)$.
Furthermore, we consider time constraints in multi-task RL. Thus, a set of MDPs is represented as 
\begin{equation}
    \{\mathcal{M}_{i, c} = (\StateSet, \ActionSet, \Dynamics_{i, c}, \RewardFunc_{i, c}, \gamma )\}_{i, c}
\end{equation}
where $c \in C$ is a time constraint.
In this multi-task RL with time constraints, an embodied agent is learned to not only satisfy the time constraints but also maximize the cumulative discounted rewards for devices $D$ where it can be deployed. 
Accordingly, the learning objective is to find the optimal policy $\pi^*$ such as
\begin{equation}
     \argmax_\pi \expectation_{I \times C} \left[ \sum_{t=0}^N \gamma^t R_{i, c}(s_t, \pi(s_t)) \right], T_D(\pi) \leq c
\end{equation}
where $T_D(\pi)$ represents the inference time on a device $D$.

\subsection{Overall Approach}
As illustrated in Figure~\ref{fig:Fig1}, we address the problem of multi-task RL with time constraints for embodied control, by employing the dynamic module selection in a multi-task modular network. Our framework $\oursol$ includes three components: a modular base network for learning diverse tasks, a module selection network for performing adaptive inference under given time constraints, and a device adapter for configuring the module utilization according to the resource availability of a specific target device.  

To achieve an adaptive policy, our approach utilizes a modular base network structure. This structure is flexible, allowing for direct adjustment of the computational load (in FLOPs) through selective module activation. It empowers the system to effectively balance the trade-off between accuracy and inference time (delay), thus enabling the time-sensitive robust model inference. To implement this structure, we adopt the soft modularization technique~\cite{yang2020multi}, which dynamically determines the weight of the path between learning modules for given task information. 
We implement the module selection network to determine the modules of the base network for each inference. This facilitates instance-wise computational adaptability, by taking task information and module utilization as input. 
By joint learning with the base network in a multi-task environment, the module selection network learns to determine the effective combination of modules for a specific task under the accuracy and inference time trade-off.
Finally, to adapt to time constraints for each device, the device adapter converts the constraints into tolerable module utilization. This allows $\oursol$ to directly use the constraints for inference. Each device adapter is tailored for its own target device through few-shot learning. 

During the model deployment on a specific target device, the device adapter infers the appropriate module utilization that adheres to given constraints. Then, the module utilization is used as input to the module selection network that determines the modules to use for each input instance (i.e., each visual state for a multi-task embodied RL agent). Individual instances contain different task information, and each can be combined with the module utilization so as to make effective decisions on the module selection.

\section{Learning A Modular Network}
We describe the joint learning procedure for the modular base network and the module selection network. 
\begin{figure*}[t] 
    \centering
    \includegraphics[width=0.92\textwidth]{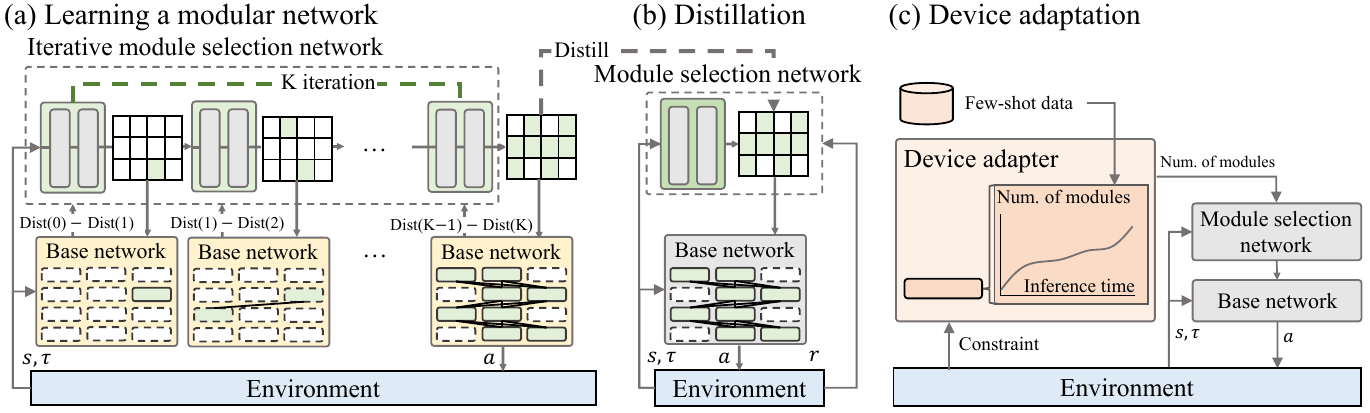}
    \caption{Learning Procedure of $\oursol$. On the left side of the figure, the base network and the iterative module selection network are jointly leaned through a reward function $R_\text{ims}$. The iterative module selection network then distills into a single-step decision module selection network, as shown in the middle side. Finally, as depicted on the right side, the device adapter utilizes few-shot samples to associate the inference time with the number of modules (module utilization), effectively transforming the constraint representation into a specific number of modules to use for different devices.} 
    \label{fig:Fig2}
\end{figure*}

\noindent \textbf{Modular base network.} 
To achieve a multi-task RL model, we employ soft modularization~\cite{yang2020multi}, a composite structure with a modular network and a soft routing network. 
The modular network infers actions based on state $s$, and the soft routing network infers the weights of paths in the modular network based on both state $s$ and task index $\tau$.
We add a module index set to use $m$ as input to the base network. 
The base network uses only the modules specified in $m$ at inference.
For batch $\mathcal{B} = \{(s, a, \tau)_i\}_{i\leq n}$ from replay buffer $\mathcal{D}_\text{base}$, we obtain a pre-trained base network $\PTBaseNet$ by optimizing multi-task loss $\mathcal{L}_\text{MTRL}$ defined as
\begin{equation}
    \begin{split}
        &\mathbb{E}_{\mathcal{B}}\left[{w_{\tau} * (\alpha_\tau \log \PTBaseNet(a|s, \tau, m_\text{full}) - Q(s, a))}\right].
    \end{split}
\label{eq:sac_loss}
\end{equation}
$\alpha_\tau$ is a temperature parameter of entropy for each task $\tau$, $m_{\text{full}}=[1, 1, 1, ..., 1]$ represents selected modules, and $Q$ is the learned Q-function.
To promote multi-task RL in the set of tasks, we adjust the learning speed for each task by adjusting the scale of loss using weights calculated as
\begin{equation}
    w_\tau = \frac{\textrm{exp}(-\alpha_\tau)}{\sum_{\tau \in \mathcal{T}} \textrm{exp}(-\alpha_\tau )}.
\end{equation}

\noindent \textbf{Joint learning.}
To enable an embodied agent to quickly adapt to time constraints, we jointly optimize the pre-trained base network and the module selection network. 
The module selection network infers module selection $m$ based on the state $s$, task-specific information $\tau$, and the number of modules to use $K$. 
The combination of the base network and module selection network can adjust the inference time by taking the number of modules to use as input and partially activating the modules of the base network.

\begin{algorithm}[t]
\begin{algorithmic}
\fontsize{9pt}{11pt}\selectfont
\STATE {Multi-task environment $env$, Replay buffer $\mathcal{D}_{\text{base}}, \mathcal{D}_{\text{ims}} = \emptyset$}
\STATE {Number of modules in base network $N$, timesteps $t$}
\STATE {Learning rate $\lambda_{\text{base}}, \lambda_{\text{ims}}$, Pre-trained base network $\bar{\pi}_\text{base}$}
\STATE {Base network $\pi_\text{base}$, Iterative module selection network $\pi_\text{ims}$}
\LOOP
\STATE {$t = 0$, $s_0, \tau_0 = env.\text{reset}()$, $\BaseNet \gets \PTBaseNet$}
\STATE {$K \sim \text{Uniform}(\{1, ..., N\})$}
\LOOP
    \STATE{$m_{0: 0} = \left[0, ... ,0\right]$}
    \FOR{$i = 1, ... ,K$}
        \STATE {$\hat{m}_{i} = \pi_{\text{ims}}(s_t, \tau_t, K, m_{0:i-1})$}
        \STATE {$m_{0:i} = m_{0:i-1} + \hat{m}_{i}$}
        \STATE {$r_i = \RewardFunc_{\text{ims}}(\BaseNet, s_t, \tau_t, m_{0:i})$ using \eqref{eq:ims_reward}}
        \STATE {$\mathcal{D}_{\text{ims}} \gets \mathcal{D}_{\text{ims}} \cup \{((s_t, \tau_t, K, m_{0:i-1}), \hat{m}_i, r_i)\}$}
    \ENDFOR
    \STATE {$\pi_{\text{ims}} \gets \pi_{\text{ims}} - \lambda_{\text{ims}}\cdot\nabla \mathcal{L}_{\text{IMS}}$ using~\eqref{eq:ims_loss}}
    
    \STATE {$a_t = \pi_{\text{base}}(s_t, \tau_t, m_{0:K})$}
    \STATE {$s_{t+1}, \tau_{t+1}, r_t =  env.\text{step}(a_t)$}
    \STATE {$\mathcal{D}_{\text{base}} \gets \mathcal{D}_{\text{base}} \cup \{(s_t, a_t, r_t, s_{t+1})\}$}
    \STATE {$\pi_{\text{base}} \gets \pi_{\text{base}} - \lambda_{\text{base}}\cdot\nabla (\mathcal{L}_{\text{MTRL}} + \mathcal{L}_{\text{RG}})$ using~\eqref{eq:sac_loss}, ~\eqref{eq:regular_loss}}
    \STATE{$t=t+1$}

\ENDLOOP
\ENDLOOP
\end{algorithmic}
\caption{Joint learning procedure}
\label{algo:train_module_selection}
\end{algorithm}

In the process of joint learning, two problems arise: (1) the combinatorial optimization problem specific to module selection, and (2) the non-stationarity in concurrent RL training. 
The former comes from an exponentially large action space in module selection, significantly degrading the performance and learning efficiency. For instance, with 16 modules, potential combinations reach approximately $10^{5}$, complicating RL exploration and increasing sample amounts~\cite{metz2017discrete}.
Furthermore, in joint learning, the interaction between the base network and the module selection network leads to a non-stationary learning environment~\cite{omidshafiei2017deep}. 

\noindent \textbf{Module selection network.} 
To address the combinatorial optimization problem, we incorporate an iterative decision making procedure into module selection. 
The iterative module selection network, denoted as $\IterMSNet$, operates by sequentially choosing an individual module across a total of $K$ iterations. This achieves the selection of $K$ modules represented as a binary vector $m \in \{0, 1\}^N$, where $1$ indicates a selected module and $0$ denotes a non-selected one. $\IterMSNet$ maps $s$, $\tau$, and the cumulative sum of module selections $m_{0:i-1}$ until the $(i-1)^{th}$ iteration to the $i^{th}$ individual module selection denoted as $\hat{m}_i \in \{0, 1\}^N$. Thus, the module selection $m$ for K modules is inferred by
\begin{equation}
    m = \sum_{i=1}^{K} \IterMSNet(s, \tau, K, m_{0:i-1}).
\label{eq:ims_inference}
\end{equation}
To train $\IterMSNet$, we directly evaluate each selection, leveraging a reward function based on the similarity in actions inferred by the base network $\BaseNet$. 

Given an action inferred through utilizing the entire modules, the reward function $\RewardFunc_{\text{ims}}$ is defined from the difference in distance between $\BaseNet(s, \tau, m_{\text{full}})$ and $\BaseNet(s, \tau, m_{0: i})$ subsequent to the previous module selection step:
\begin{equation}
    \RewardFunc_{\text{ims}}(\BaseNet, s, \tau, m_{0:i}) = \text{Dist}(i-1) - \text{Dist}(i)
\label{eq:ims_reward}
\end{equation}
where $\text{Dist}(i) = ||\BaseNet(s, \tau, m_{\text{full}}) - \BaseNet(s, \tau, m_{0:i})||$.
This reward function not only accelerates the learning of the iterative module selection network but also minimizes the regret bounds of the actions generated by the base network.
When representing $\RewardFunc$ for given task $\tau$ as an $L$-Lipschitz function, we can obtain the upper bound of the difference of rewards in a multi-task environment.
\begin{equation}
    \begin{aligned}
    |\RewardFunc(s, \BaseNet(s, \tau, m_{\text{full}}))  - \RewardFunc(s, \BaseNet&(s, \tau, m_{0:i})) | \\
        & \leq L \cdot \text{Dist}(i)
    \end{aligned}
\label{eq:ims_obj}
\end{equation}
Thus, by minimizing $\text{Dist}(K)$, the difference in rewards in Eq.~\eqref{eq:ims_reward} is also minimized. This allows the actions inferred using a subset of modules to closely approximate the optimal reward.

To mitigate performance drops caused by the non-stationary problems, we avoid dramatic changes in actions between the pre-trained base network and the fine-tuned base network by using a regularization loss. 
Let $\mathcal{B} =\{((s, \tau, K, m_{0:t_i-1}), \hat{m}_{t_i}, r)_i\}_{i\leq n}$ be a sample batch from replay buffer $\mathcal{D}_\text{ims}$. The regularization loss $\mathcal{L}_{\text{RG}}$ is defined as
\begin{equation}
    \mathbb{E}_{\mathcal{B}}\left[\lVert \bar{\pi}_{\text{base}}(s, \tau, m_{\text{full}}) - \pi_{\text{base}}(s, \tau, m_{0:t_i})\rVert\right]
\label{eq:regular_loss}
\end{equation}
where $\bar{\pi}_{\text{base}}$ is the pre-trained base network without additional learning. The module selection $m_{\text{full}}$ implies that whole modules of the base network are selected.
Furthermore, we combine Reptile~\cite{nichol2018first}, a meta-RL algorithm, with REINFORCE~\cite{williams1992simple}.
The policy loss for the iterative module selection network $\mathcal{L}_{\text{IMS}}$ is defined as
\begin{equation}
    \mathbb{E}_{\mathcal{B}}\left[\sum_{i=1}^n \log \pi_{\text{ims}}(\hat{m}_{t_i} | s_i, \tau, K, m_{0:t_i-1}) \cdot G_i\right].
\label{eq:ims_loss}
\end{equation}
Here, $G_i$ is the discounted cumulative sum of rewards at timesteps $i$. 
Given that $\pi_{\text{ims}}^k$ is the result of $k$ updates from $\pi_{\text{ims}}$ with the sample batch relative to the current $\pi_{\text{base}}$, the update for $\pi_{\text{ims}}$ is executed with the step-size parameter $\epsilon$.
\begin{equation}
    \pi_{\text{ims}} \gets \pi_{\text{ims}} + \epsilon(\pi_{\text{ims}}^k - \pi_{\text{ims}})
\end{equation}

The joint learning procedure of the base network and the iterative module selection network is illustrated on the left side of Figure~\ref{fig:Fig2}, with details provided in Algorithm~\ref{algo:train_module_selection}.

\begin{algorithm}[t]
\begin{algorithmic}
\fontsize{9pt}{11pt}\selectfont
\STATE {Multi-task environment $env$, device $D$, time-constraint $c$}
\STATE {Number of modules in base network $N$, timesteps $t$}
\STATE {Base network $\pi_\text{base}$, Iterative module selection network $\pi_\text{ims}$}
\STATE {Module selection network $\pi_\text{ms}$, Device adapter $\pi_\text{da}$}
\STATE {Inference time $T_D$ for device $D$}
\STATE {Dataset $\mathcal{D}_\text{ms}, \mathcal{D}_\text{da} = \emptyset$, Learning rate $\lambda_{\text{ms}}$, $\lambda_{\text{da}}$}

\color{blue}
\STATE{\textit{/* Distillation-based optimization of $\MSNet$ */}}
\color{black}

\LOOP
\STATE {$t=0$, $s_0, \tau_0=env.reset()$}
\STATE {$K \sim \text{Uniform}(\{1, ..., N\})$}
\LOOP
    \STATE {$m_t = \pi_{\text{ms}}(s_t, \tau_t, K)$,\ $a_t = \pi_{\text{base}}(s_t, \tau_t, m_t)$}
    \STATE {$m_{0:K} = \sum^K_{i=0} \pi_{\text{ims}}(s_t, \tau_t, K, m_{0:i})$}
    \STATE {$s_{t+1}, \tau_{t+1}, r_t =  env.step(a)$}
    \STATE {$r_t = r_t - ||m_t - m_{0:K}||_2$}
    \STATE {$\mathcal{D}_{\text{ms}} \gets \mathcal{D}_{\text{ms}} \cup \{((s_t, \tau_t, K), m_t, r_t, (s_{t+1}, \tau_{t+1}, K))\}$}
    \STATE {$\pi_{\text{ms}} \gets \pi_{\text{ms}} - \lambda_{\text{ms}}\cdot\nabla (\mathcal{L}_{\text{MTRL}} + \mathcal{L}_{\text{KD}})$ using~\eqref{eq:sac_loss},~\eqref{equ: distill_loss}}
    \STATE {$t=t+1$}
\ENDLOOP
\ENDLOOP

\color{blue}
\STATE{\textit{/* Few-shot adaptation through device adapter $\ModeLearner$ */}}
\color{black}
\LOOP
\STATE {$t = 0, s_0, \tau_0 = env.\text{reset}()$}
\STATE {$K \sim \text{Uniform}(\{1, ..., N\})$}
\LOOP
    \STATE {$c_t, a_t = T_D(\BaseNet(s_t, \tau_t, \MSNet(s_t, \tau_t, K)))$}
    \STATE {$s_{t+1}, \tau_{t+1}, r_t =  env.\text{step}(a_t)$}
    \STATE {$\mathcal{D}_\text{da} \gets \mathcal{D}_\text{da} \cup \{(K, c_t)\}$}
    \STATE {$\ModeLearner \gets \ModeLearner - \lambda_{\text{da}}\cdot\nabla  \mathcal{L}_\text{DA}$} using ~\eqref{equ: adpative_loss}
    \STATE {$t=t+1$}
\ENDLOOP
\ENDLOOP

\end{algorithmic}
\caption{Distillation-based optimization}
\label{algo:PGTE}
\end{algorithm}

\section{Distillation-based Optimization}
We describe two schemes tailored for device-specific adaptation, the knowledge distillation for the module selection network and the few-shot learning for the device adapter. 
To enhance the efficiency of the module selection network, we reconstruct it with single-step inference through knowledge distillation. While the iterative module selection network shows superior performance in the large action space, its inference often incurs excessive delays and computational loads compared to the base network. 
The module selection network, denoted as $\pi_\text{ms}$, takes a $\mathcal{B} = \{(s, \tau, K)_i\}_{i<n}$ from replay buffer $\mathcal{D}_\text{ms}$ as input in a single step. To train $\pi_{\text{ms}}$ based on $\pi_{\text{ims}}$, we use $\mathcal{L}_\text{MTRL}$ in Eq.~\eqref{eq:sac_loss}, with the knowledge distillation loss $\mathcal{L}_\text{KD}$ defined as
\begin{equation}
    \begin{aligned}
	\mathbb{E}_{\mathcal{B}}\left[\lVert \pi_{\text{ms}}(s, \tau, K) -  \sum^K_{i=1}\pi_{\text{ims}}(s, \tau, K, m_{0:i-1}) \rVert \right].
    \end{aligned}
	\label{equ: distill_loss}
\end{equation}
This distillation not only ensures the maintenance of the module selection performance but also considerably reduces the inference time with the module selection network.

\begin{table*}[t]
    \caption{Performance for Meta-world single task in the success rate with 95\% confidence intervals: the best performance is in bold.}
    \footnotesize
    \centering
    \begin{adjustbox}{width=0.96\textwidth}
        \begin{tabular}{|c|c||c|c|c|c|c|c|c|c|c|c|}
            \hline
             \multirow{2}{*}{Device} & \multirow{2}{*}{Constraint} & \multicolumn{2}{c|}{DRNet} & \multicolumn{2}{c|}{D2NN} & \multicolumn{2}{c|}{DS-Net} & \multicolumn{2}{c|}{RL-AA} & \multicolumn{2}{c|}{\textoursol} \\
             \cline{3-12}
             \ & & Success rate & FLOPs & Success rate & FLOPs & Success rate & FLOPs & Success rate & FLOPs & Success rate & FLOPs \\
            \hline
            \multirow{5}{*}{Orin} & 8 $ms$ & - & - & $26.7 \pm 3.9\%$ & 62M & $34.8 \pm 11.7\%$ & 25M & $35.4 \pm 4.4\%$ & 149M & $\mathbf{57.5 \pm 9.9\%}$ & 92M \\
             & 10 $ms$ & $18.0 \pm 4.5\%$ & 190M & $27.3 \pm 3.5\%$ & 62M & $30.5 \pm 9.3\%$ & 159M & $34.9 \pm 5.0\%$ & 149M & $\mathbf{65.0 \pm 13.4\%}$ & 151M \\
             & 12 $ms$ & $18.0 \pm 4.5\%$ & 190M & $32.0 \pm 4.4\%$ & 110M & $31.8 \pm 12.7\%$ & 403M & $41.2 \pm 4.1\%$ & 355M & $\mathbf{75.0 \pm 9.4\%}$ & 229M \\
             & 14 $ms$ & $38.1 \pm 10.6\%$ & 364M & $54.0 \pm 6.9\%$ & 176M & $31.5 \pm 9.9\%$ & 567M & $47.1 \pm 5.0\%$ & 360M & $\mathbf{74.5 \pm 14.2\%}$ & 254M \\
             & 16 $ms$ & $39.8 \pm 11.7\%$ & 491M & $68.0 \pm 8.9\%$ & 284M & $30.0 \pm 8.1\%$ & 758M & $47.3 \pm 5.9\%$ & 365M & $\mathbf{74.5 \pm 14.2\%}$ & 254M \\
            \hline
            \multirow{5}{*}{Xavier} & 12 $ms$ & - & - & $27.3 \pm 4.2\%$ & 62M & $31.0 \pm 9.5\%$ & 25.48M & $33.4 \pm 5.3\%$ & 149M & $\mathbf{50.0 \pm 7.5\%}$ & 92M  \\
             & 15 $ms$ & $18.0 \pm 4.5\%$ & 190M & $36.0 \pm 6.8\%$ & 62M & $30.0 \pm 8.5\%$ & 78M & $32.1 \pm 4.9\%$ & 205M & $\mathbf{56.0 \pm 9.4\%}$ & 151M \\
             & 18 $ms$ & $20.0 \pm 6.7\%$ & 243M & $38.2 \pm 9.0\%$ & 62M & $26.0 \pm 6.9\%$ & 159M & $41.3 \pm 4.4\%$ & 355M & $\mathbf{86.0 \pm 9.7\%}$ & 214M \\
             & 21 $ms$ & $20.0 \pm 6.7\%$ & 243M & $33.8 \pm 5.9\%$ & 160M & $29.3 \pm 8.8\%$ & 403M & $45.7 \pm 8.5\%$ & 365M & $\mathbf{80.0 \pm 13.5\%}$ & 254M \\
             & 24 $ms$ & $38.0 \pm 10.7\%$ & 391M & $69.6 \pm 9.0\%$ & 284M & $28.0 \pm 7.4\%$ & 567M & $48.0 \pm 6.4\%$ & 365M & $\mathbf{80.0 \pm 13.5\%}$ & 254M \\
            \hline
            \multirow{5}{*}{Nano} & 40 $ms$ & - & - & $30.0 \pm 7.5\%$ & 62M & $32.0 \pm 6.7\%$ & 78M & $32.1 \pm 6.1\%$ & 149M & $\mathbf{40.0 \pm 9.2\%}$ & 114M  \\
             & 46 $ms$ & $18.0 \pm 4.5\%$ & 190M & $24.1 \pm 6.0\%$ & 82M & $30.7 \pm 4.0\%$ & 159M & $40.4 \pm 7.7\%$ & 355M & $\mathbf{47.5 \pm 13.2\%}$ & 158M \\
             & 52 $ms$ & $37.9 \pm  10.5\%$ & 320M & $38.0 \pm 12.5\%$ & 82M & $28.7 \pm 3.9\%$ & 403M & $42.3 \pm 5.3\%$ & 355M & $\mathbf{72.5 \pm 10.2\%}$ & 214M \\
             & 58 $ms$ & $40.9 \pm 10.3\%$ & 485M & $38.7 \pm 10.3\%$ & 180M & $28.7 \pm 4.5\%$ & 567M & $46.1 \pm 4.9\%$ & 365M & $\mathbf{65.0 \pm 9.2\%}$ & 258M \\
             & 64 $ms$ & $42.0 \pm 12.5\%$ & 539M & $\mathbf{68.7 \pm 5.5\%}$ & 269M & $26.7 \pm 3.9\%$ & 758M & $46.9 \pm 4.4\%$ & 365M & $65.0 \pm 9.2\%$ & 258M \\
            \hline
        \end{tabular}
    \end{adjustbox}
    \label{tab:meta_single}
\end{table*}

\section{Few-Shot Device Adaptation}
To make our model adaptable under various time constraints for a specified device, we employ a device adapter that can manipulate the inference of the module selection network.
This device adapter takes a time constraint, determining the number of modules to use, ensuring that it does not violate the constraint. By using the adapter, time constraints are directly grounded as values within the network, enabling $\oursol$ to perform the constraint-aware inference.

To train the device adapter $\ModeLearner$, we use a pre-trained base network $\BaseNet$ and distilled module selection model $\MSNet$ with a loss function specifically designed to accommodate the constraints of the current device. 
For a given device, we collect a model inference dataset $\mathcal{D}_\textbf{da}$ and sample batch denoted as $\mathcal{B}=\{(K, c)_i\}_{i<n}$, where $K$ is the number of modules to use and $c$ is the inference time of $\oursol$ when using only $K$ modules. The device adapter is optimized by $\mathcal{L}_\text{DA}$.
\begin{equation}
    \begin{aligned}
	\mathbb{E}_{\mathcal{B}}\left[ \lVert \ModeLearner(c) - K \rVert \cdot (1 - p) \right]
    \end{aligned}
	\label{equ: adpative_loss}
\end{equation}
Here, $p$ represents the penalty weights applied when the device adapter predicts the number of modules exceeding $K$; otherwise, $p$ is set to 0. 

The procedure of distillation-based optimization and few-shot device adaptation is illustrated on the right side of Figure~\ref{fig:Fig2}, with details provided in Algorithm~\ref{algo:PGTE}.

\section{Evaluation}
\subsection{Environments and Devices}
\begin{figure}[htbp] 
    \centering
    \begin{subfigure}{0.32\linewidth}
        \centering
        \includegraphics[width=\linewidth]{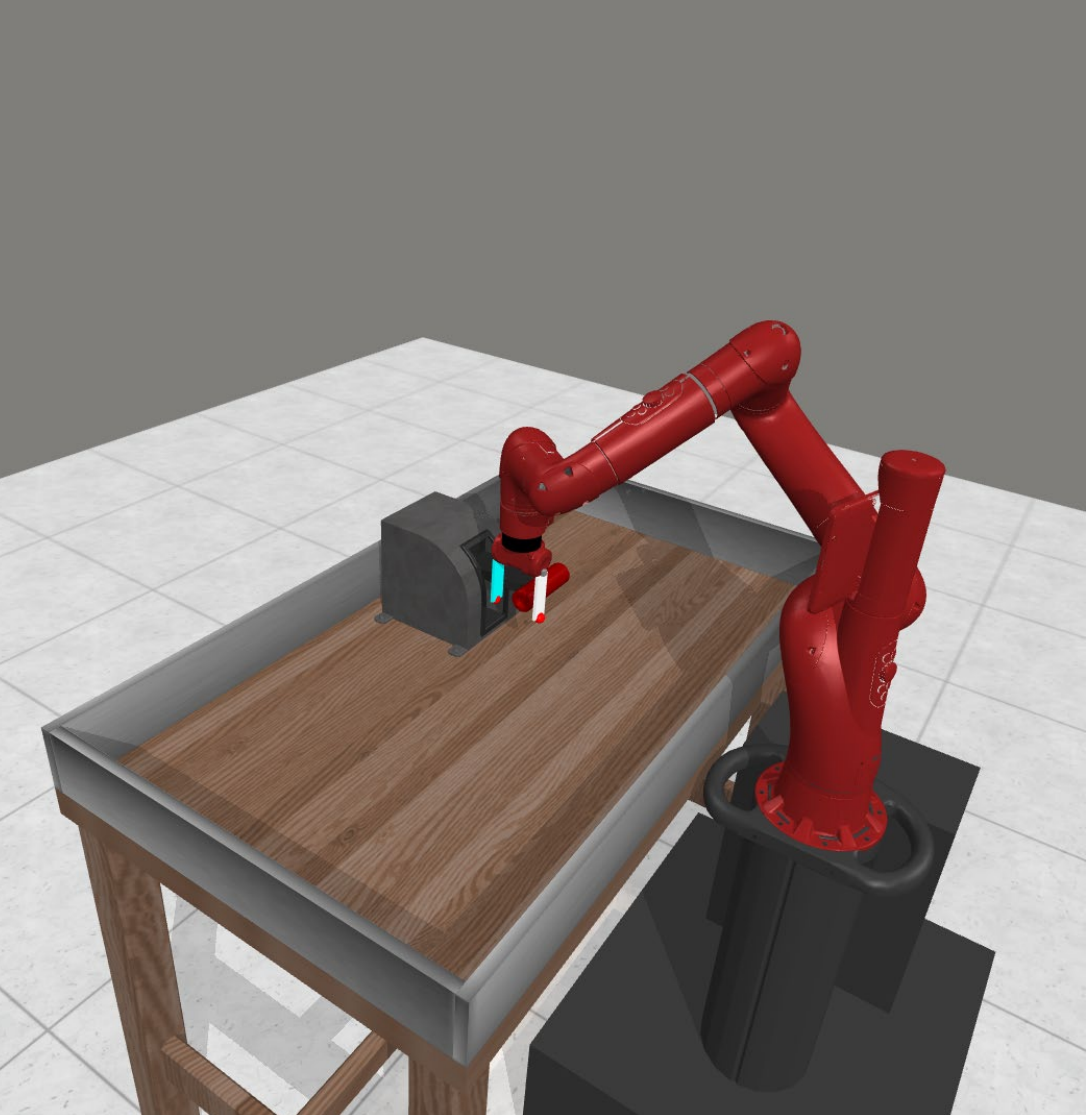} 
        \caption{Meta-world}
    \end{subfigure}
    \begin{subfigure}{0.32\linewidth}
        \centering
        \includegraphics[width=\linewidth]{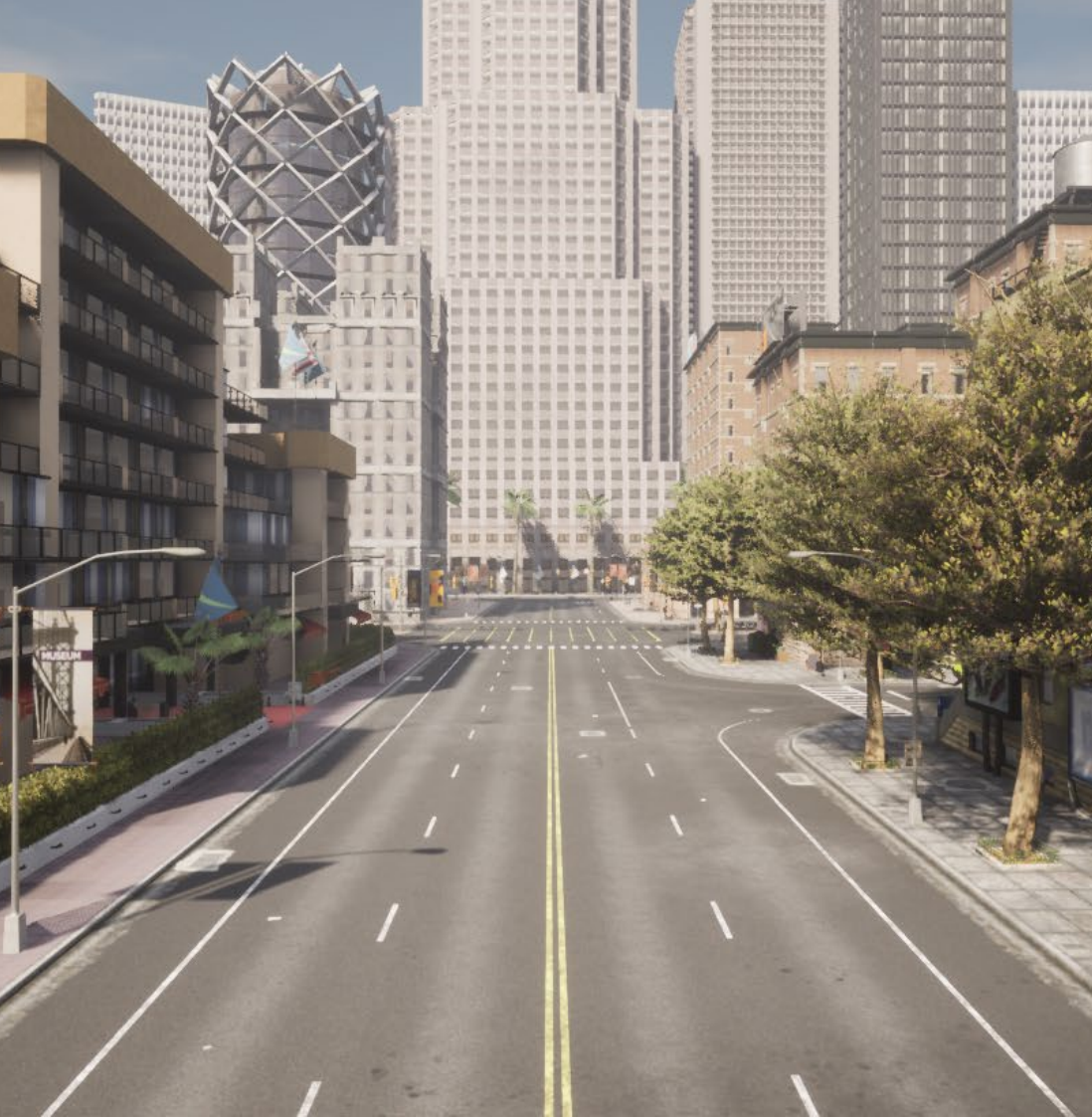} 
        \caption{CARLA}
    \end{subfigure}
    \begin{subfigure}{0.32\linewidth}
        \centering
        \includegraphics[width=\linewidth]{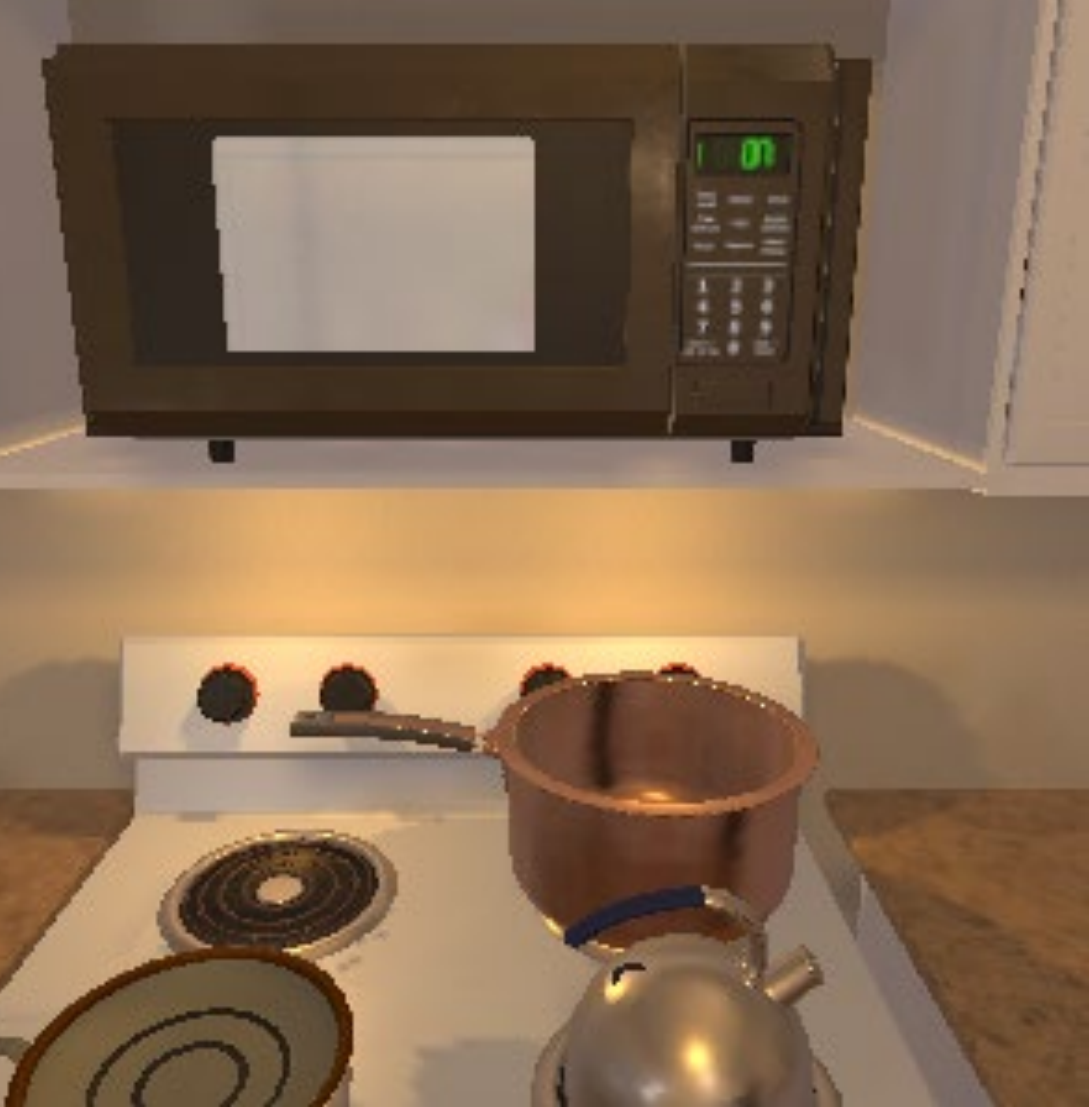} 
        \caption{AI2THOR}
    \end{subfigure}
    \caption{Environments}
    \label{fig:env}
    \vspace{-0.08in}
\end{figure}

\noindent \textbf{Meta-world}.
We use the MT10 benchmark (i.e., 10 different control tasks) in Meta-world~\cite{yu2020meta}, where each task is given a specific manipulation objective such as opening a door or closing a window. We compare the performance of robot manipulation tasks under time constraints.

\noindent \textbf{CARLA.}
To demonstrate mission-critical scenarios where the inference time is of critical importance, we use the autonomous driving simulator CARLA~\cite{dosovitskiy2017carla}. Models are trained for autonomous driving tasks with vision-based states at a multi-task configuration with 12 different maps. 

\noindent \textbf{AI2THOR.} 
We use AI2THOR~\cite{kolve2017ai2}, 
where an agent navigates the map with egocentric vision states, placing various objects to complete a rearrangement task. The simulation environments are represented in Figure~\ref{fig:env}.

In our evaluation, we test several embedded devices, each with distinct resources and computational capabilities. The devices include Nvidia Jetson Nano (Nano), Nvidia Jetson Xavier NX 8GB (Xavier), and Nvidia Jetson AGX Orin 32GB (Orin), with the Nano being the least powerful, followed by the Xavier and the Orin being the most powerful. 
By testing on these devices with varying levels of capabilities, we can better understand how our framework adapts to different resource limitations. This is crucial in embodied AI, where deployment environments can greatly vary in terms of available computational resources. 
The detailed device specifications are in Table~\ref{tab:devices}.

\begin{table}[h!]
    \caption{Device Specification}
    \begin{adjustbox}{width=0.47\textwidth}
    \footnotesize
    \centering
        \begin{tabular}{|c||c|c|c|c|}
            \hline
             Device & Performance & CPU Max Freq. & GPU Max Freq. & Memory \\
            \hline
            Orin & 275 TOPs & 2.2 GHz & 1.3 GHz & 32 GBs \\
            \hline
            Xavier & 21 TOPs & 1.9 GHz & 1.1 GHz & 8 GBs \\
            \hline
            Nano & 472 GFLOPs & 0.9 GHz & 0.6 GHz & 4 GBs \\
            \hline
        \end{tabular}
    \end{adjustbox}
    \label{tab:devices}
    \vspace{-0.1in}
\end{table}

\begin{table*}[t]
    \caption{Performance for Meta-world multi-task in the success rate with 95\% confidence intervals}
    \footnotesize
    \centering
    \begin{adjustbox}{width=0.96\textwidth}
        \begin{tabular}{|c|c||c|c|c|c|c|c|c|c|c|c|}
            \hline
             \multirow{2}{*}{Device} & \multirow{2}{*}{Constraint} & \multicolumn{2}{c|}{DRNet} & \multicolumn{2}{c|}{D2NN} & \multicolumn{2}{c|}{DS-Net} & \multicolumn{2}{c|}{RL-AA} & \multicolumn{2}{c|}{\textoursol} \\
             \cline{3-12}
             \ & & Success rate & FLOPs & Success rate & FLOPs & Success rate & FLOPs & Success rate & FLOPs & Success rate & FLOPs \\
            \hline
            \multirow{5}{*}{Orin} & 8 $ms$ & $-$ & - & $20.0 \pm 6.7\%$ & 92M & $11.7 \pm 7.1\%$ & 25M & $15.7 \pm 6.9\%$ & 149M & $\mathbf{27.1 \pm 8.3\%}$ & 92M  \\
             & 10 $ms$ & $12.0 \pm 4.2\%$ & 119M & $18.0 \pm 5.6\%$ & 151M & $11.6 \pm 4.4\%$ & 159M & $15.3 \pm 7.1\%$ & 149M & $\mathbf{30.3 \pm 10.1\%}$ & 151M \\
             & 12 $ms$ & $26.7 \pm 5.4\%$ & 488M & $17.0 \pm 4.8\%$ & 229M & $23.3 \pm 8.5\%$ & 403M & $18.5 \pm 8.6\%$ & 355M & $\mathbf{34.9 \pm 9.7\%}$ & 229M \\
             & 14 $ms$ & $26.7 \pm 6.1\%$ & 488M & $51.9 \pm 10.4\%$ & 254M & $16.7 \pm 8.7\%$ & 567M & $26.1 \pm 9.1\%$ & 355M & $\mathbf{53.0 \pm 6.8\%}$ & 254M \\
             & 16 $ms$ & $30.1 \pm 4.3\%$ & 636M & $\mathbf{54.1 \pm 8.4\%}$ & 254M & $25.0 \pm 5.7\%$ & 758M & $28.3 \pm 7.1\%$ & 365M & $53.0 \pm 6.8\%$ & 254M \\
            \hline
            \multirow{5}{*}{Xavier} & 12 $ms$ & $-$ & - & $20.4 \pm 6.7\%$ & 92M & $18.3 \pm 7.9\%$ & 25M & $14.0 \pm 7.8\%$ & 149M & $\mathbf{26.7 \pm 7.6\%}$ & 92M \\
             & 15 $ms$ & $10.3 \pm 3.2\%$ & 119M & $23.5 \pm 6.9\%$ & 151M & $16.7 \pm 5.4\%$ & 78M & $14.9 \pm 5.3\%$ & 149M & $\mathbf{29.5 \pm 10.1\%}$ & 151M \\
             & 18 $ms$ & $19.8 \pm 4.5\%$ & 488M & $17.3 \pm 4.8\%$ & 214M & $18.3 \pm 4.2\%$ & 159M & $23.1 \pm 9.2\%$ & 355M & $\mathbf{28.1 \pm 9.1\%}$ & 214M \\
             & 21 $ms$ & $18.8 \pm 3.7\%$ & 488M & $56.8 \pm 10.1\%$ & 254M & $21.7 \pm 10.3\%$ & 403M & $28.1 \pm 6.2\%$ & 365M & $\mathbf{53.6 \pm 5.9\%}$ & 254M \\
             & 24 $ms$ & $21.2 \pm 3.8\%$ & 636M & $53.0 \pm 9.0\%$ & 254M & $15.0 \pm 5.8\%$ & 567M & $27.6 \pm 6.5\%$ & 365M & $\mathbf{53.6 \pm 5.9\%}$ & 254M \\
            \hline
            \multirow{5}{*}{Nano} & 40 $ms$ & $-$ & - & $23.5 \pm 15.2\%$ & 114M & $10.4 \pm 8.9\%$ & 78M & $16.1 \pm 6.1\%$ & 149M & $\mathbf{27.0 \pm 7.5\%}$ & 114M \\
             & 46 $ms$ & $10.0 \pm 0.8\%$ & 119M & $17.5 \pm 8.0\%$ & 158M & $9.9 \pm 2.1\%$ & 159M & $\mathbf{23.4 \pm 7.0\%}$ & 355M & $23.0 \pm 5.8\%$ & 162M \\
             & 52 $ms$ & $23.3 \pm 5.4\%$ & 488M & $20.0 \pm 18.3\%$ & 214M & $13.3 \pm 5.3\%$ & 267M & $24.1 \pm 6.5\%$ & 365M & $\mathbf{28.1 \pm 8.1\%}$ & 217M \\
             & 58 $ms$ & $16.7 \pm 4.9\%$ & 488M & $52.3 \pm 8.0\%$ & 258M & $20.0 \pm 9.4\%$ & 567M & $27.4 \pm 8.6\%$ & 365M & $\mathbf{53.9 \pm 6.0\%}$ & 234M \\
             & 64 $ms$ & $23.3 \pm 6.1\%$ & 636M & $52.5 \pm 17.9\%$ & 258M & $20.0 \pm 5.4\%$ & 758M & $26.9 \pm 6.9\%$ & 365M & $\mathbf{62.1 \pm 4.5\%}$ & 254M \\
            \hline
        \end{tabular}
    \end{adjustbox}
    \label{tab:meta_multi}
\end{table*}

\begin{table*}[t]
    \caption{Performance on CARLA in the success rate with 95\% confidence intervals}
    \footnotesize
    \centering
    \begin{adjustbox}{width=0.96\textwidth}
        \begin{tabular}{|c|c||c|c|c|c|c|c|c|c|c|c|c|}
            \hline
             \multirow{2}{*}{Device} & \multirow{2}{*}{Constraint} & \multicolumn{2}{c|}{DRNet} & \multicolumn{2}{c|}{D2NN} & \multicolumn{2}{c|}{DS-Net} & \multicolumn{2}{c|}{RL-AA} & \multicolumn{2}{c|}{\textoursol} \\
             \cline{3-12}
             \ & & Success rate & FLOPs & Success rate & FLOPs & Success rate & FLOPs & Success rate & FLOPs & Success rate & FLOPs \\
            \hline
            \multirow{5}{*}{Orin} & 8 $ms$ & $0.0 \pm 0.0\%$ & 16M & $0.0 \pm 0.0\%$ & 32M & $0.0 \pm 0.0\%$ & 3M & $0.0 \pm 0.0\%$ & 34M & $\mathbf{8.3 \pm 5.0\%}$ & 31M \\
             & 10 $ms$ & $0.0 \pm 0.0\%$ & 16M & $0.0 \pm 0.0\%$ & 36M & $39.9 \pm 9.0\%$ & 10M & $0.0 \pm 0.0\%$ & 34M & $\mathbf{49.7 \pm 9.1\%}$ & 47M \\
             & 12 $ms$ & $70.1 \pm 6.1\%$ & 27M & $16.6 \pm 6.8\%$ & 42M & $71.1 \pm 7.2\%$ & 36M & $33.1 \pm 5.9\%$ & 42M & $\mathbf{83.1 \pm 6.8\%}$ & 60M \\
             & 14 $ms$ & $68.2 \pm 5.4\%$ & 48M & $41.6 \pm 8.9\%$ & 57M & $66.7 \pm 8.6\%$ & 76M & $75.6 \pm 4.9\%$ & 72M & $\mathbf{79.2 \pm 6.7\%}$ & 73M \\
             & 16 $ms$ & $70.3 \pm 5.6\%$ & 48M & $83.3 \pm 6.7\%$ & 72M & $76.7 \pm 7.7\%$ & 132M & $77.1 \pm 5.1\%$ & 72M  & $\mathbf{85.8 \pm 5.8\%}$ & 73M \\
            \hline
            \multirow{5}{*}{Xavier} & 12 $ms$ & $0.0 \pm 0.0\%$ & 16M & $0.0 \pm 0.0\%$ & 36M & $0.0 \pm 0.0\%$ & 3M & $0.0 \pm 0.0\%$ & 34M & $\mathbf{12.3 \pm 6.9\%}$ & 35M \\
             & 15 $ms$ & $0.0 \pm 0.0\%$ & 16M & $0.0 \pm 0.0\%$ & 36M & $35.5 \pm 5.0\%$ & 10M & $0.0 \pm 0.0\%$ & 34M & $\mathbf{41.7 \pm 7.7\%}$ & 47M \\
             & 18 $ms$ & $52.1 \pm 13.1\%$ & 27M & $27.8 \pm 8.6\%$ & 48M & $77.1 \pm 9.1\%$ & 36M & $36.1 \pm 5.1\%$ & 42M & $\mathbf{80.8 \pm 6.7\%}$ & 59M \\
             & 21 $ms$ & $71.1 \pm 7.3\%$ & 48M & $41.7 \pm 9.4\%$ & 54M & $75.2 \pm 7.9\%$ & 76M & $58.1 \pm 4.8\%$ & 58M & $\mathbf{79.1 \pm 7.3\%}$ & 71M \\
             & 24 $ms$ & $72.2 \pm 8.6\%$ & 48M & $\mathbf{89.3 \pm 4.4\%}$ & 72M & $83.3 \pm 6.8\%$ & 132M & $76.4 \pm 5.9\%$ & 72M & $83.3 \pm 6.7\%$ & 73M \\
            \hline
            \multirow{5}{*}{Nano} & 40 $ms$ & $0.0 \pm 0.0\%$ & 16M & $0.0 \pm 0.0\%$ & 16M & $0.0 \pm 0.0\%$ & 3M & $0.0 \pm 0.0\%$ & 34M & $\mathbf{10.4 \pm 4.1\%}$ & 35M \\
             & 46 $ms$ & $0.0 \pm 0.0\%$ & 16M & $0.0 \pm 0.0\%$ & 36M & $0.0 \pm 0.0\%$ & 3M & $32.7 \pm 5.4\%$ & 42M &  $\mathbf{35.5 \pm 10.3}\%$ & 43M \\
             & 52 $ms$ & $70.7 \pm 5.6\%$ & 27M & $51.2 \pm 6.6\%$ & 54M & $23.1 \pm 15.3\%$ & 10M & $34.5 \pm 4.6\%$ & 42M & $\mathbf{77.1 \pm 5.3\%}$ & 56M \\
             & 58 $ms$ & $69.1 \pm 7.1\%$ & 48M & $\mathbf{83.5 \pm 7.0\%}$ & 72M & $59.1 \pm 11.2\%$ & 32M & $74.1 \pm 4.3\%$ & 72M & $82.0 \pm 5.7\%$ & 71M \\
             & 64 $ms$ & $72.9 \pm 5.3\%$ & 48M & $\mathbf{85.8 \pm 6.1\%}$ & 72M & $65.5 \pm 6.7\%$ & 36M & $75.3 \pm 5.6\%$ & 72M & $81.9 \pm 6.1\%$ & 73M \\
            \hline
        \end{tabular}
    \end{adjustbox}
    \label{tab:CARLA}
\end{table*}

\begin{table}[t]
    \caption{Performance on AI2THOR in the success rate}
    \footnotesize
    \centering
    \begin{adjustbox}{width=0.44\textwidth}
        \begin{tabular}{|c|c||c|c|c|c|c|}
            \hline
             \multirow{2}{*}{Device} & \multirow{2}{*}{Constraint} & \multicolumn{2}{c|}{DS-Net} & \multicolumn{2}{c|}{\textoursol} \\
            \cline{3-6}
            \ & & Seen & Unseen & Seen & Unseen \\
            \hline
            \multirow{5}{*}{Orin} & 8 $ms$ & $98.7\%$ & $5.8\%$ & $97.9\%$ & $\mathbf{21.2\%}$ \\
             & 10 $ms$ & $96.3\%$ & $5.2\%$ & $98.0\%$ & $\mathbf{45.4\%}$ \\
             & 12 $ms$ & $98.5\%$ & $4.8\%$ & $97.5\%$ & $\mathbf{41.7\%}$ \\
             & 14 $ms$ & $96.3\%$ & $6.0\%$ & $98.2\%$ & $\mathbf{71.3\%}$ \\
             & 16 $ms$ & $97.5\%$ & $7.4\%$ & $97.9\%$ & $\mathbf{81.2\%}$ \\
            \hline
            \multirow{5}{*}{Xavier} & 12 $ms$ & $97.1\%$ & $6.2\%$ & $97.2\%$ & $\mathbf{20.7\%}$ \\
             & 15 $ms$ & $95.7\%$ & $8.0\%$ & $98.7\%$ & $\mathbf{44.0\%}$ \\
             & 18 $ms$ & $98.1\%$ & $7.4\%$ & $97.5\%$ & $\mathbf{45.2\%}$ \\
             & 21 $ms$ & $97.7\%$ & $7.0\%$ & $98.0\%$ & $\mathbf{69.3\%}$ \\
             & 24 $ms$ & $98.0\%$ & $8.2\%$ & $98.1\%$ & $\mathbf{79.1\%}$ \\
            \hline
        \end{tabular}
    \end{adjustbox}
    \label{tab:ai2thor}
\end{table}

\subsection{Comparisons}
We use several dynamic model adaptation methods as baselines.
\begin{itemize}
\item Dynamic Routing Network (DRNet)~\cite{cai2021dynamic} is a network comprising serially connected cells, each corresponding to a directed acyclic graph of nodes. It optimizes by learning to select paths between the nodes through a loss function that balances the inference time and performance. Unlike $\oursol$ which adapts a single model to different conditions, we use individual networks specifically learned for each constraint condition. 
We consider DRNet as a baseline for dynamic (adaptive) inference models.

\item Dynamic Deep Neural Networks (D2NN)~\cite{liu2018dynamic} is a modular neural network that exploits the accuracy-efficiency trade-off. It exploits RL in module selection, using the rewards calculated according to performance and inference time. Unlike $\oursol$, we use individual networks specifically learned for each constraint condition. 
We consider D2NN as an RL baseline tailored for trade-off conditions. 

\item Dynamic Slimmable Network (DS-Net)~\cite{li2021dynamic} is a dynamic network model, in which an internal gater network determines the weight ratio for convolution neural layers to slice the network for inference. 
To align with our problem formulation, we modify the gater network (taking the ratio as input) in a way of conducting instance-wise dynamic inference, similar to our approach. In our comparison, DS-Net serves as a baseline for adaptive models that handle multiple constraints within a single policy.

\item RL via Asymmetric Architecture (RL-AA)~\cite{chang2021reducing} is a hierarchical policy to dynamically adjust the module usage. The low-level policy consists of two models, each with a small and large scale, and the high-level policy determines which policy to use. 
To adapt to various time-constrained conditions, we include a wider range of low-level policies, each with varying inference time. We use RL-AA as a baseline for resource-adaptive RL methods.
\end{itemize}

\subsection{Adaptation Performance}
\textbf{Meta-world Single-task.}
For 5 individual tasks in MT10, Table~\ref{tab:meta_single} shows the performance under various time constraints (in the column of ``Constraint''), achieved by our $\oursol$ and the baselines (DRNet, D2NN, DS-Net, RL-AA). 
Specifically, we evaluate the average success ratio and computation load (in FLOPs) within the constraint violation rate of 1\% for 3 different devices (in the column of ``Constraint''). 
As shown, $\oursol$ achieves superior performance for most configurations. 
Compared to D2NN, the most competitive baseline, $\oursol$ achieves a 25.1\% gain. 
While sharing a common base network structure with D2NN, $\oursol$ shows better performance, as it employs the iterative module selection and distillation.
More importantly, D2NN needs to be retrained for each configuration (i.e., each constraint and device setting). 
$\oursol$ achieves this performance superiority across different configurations, using only a single model without retraining, demonstrating its adaptation capabilities to different time and resource constraints.

\noindent \textbf{Meta-world Multi-task.}
Table~\ref{tab:meta_multi} compares the performance of the MT10 multi-task. 
$\oursol$ demonstrates consistently its performance superiority, achieving an average performance gain of 5.7\% over D2NN, the most competitive baseline. This specifies the adaptation capabilities of $\oursol$, achieved not only through module selection but also through multi-task learning for the base network.

\noindent \textbf{CARLA.}
Table~\ref{tab:CARLA} shows the performance for autonomous driving tasks across 12 different maps in CARLA, where delayed inference often degrades the performance and poses risks; we implement such a strategy that upon a constraint violation (i.e., inference delay), the action at the previous timestep is reused. 
$\oursol$ shows 14.4\% higher performance than DS-Net, which is the most competitive comparison in this experiment. 
Due to the direct impacts of constraint violations in CARLA, the adaptive inference is more beneficial, compared to the Meta-World tasks. This leads to better performance by the methods capable of constraint-aware inference, such as ours and DS-Net.  
DS-Net shows a significant performance drop in Nano, which is a small memory device; DS-Net requires a large computation load per single layer, unlike ours.

\noindent \textbf{AI2THOR.}
Table~\ref{tab:ai2thor} compares the performance for AI2THOR's complex navigation tasks, where ``Seen" refers to initial object positions encountered during training and ``Unseen" refers to those not during training. 
Both $\oursol$ and DS-Net perform well in the seen configurations, but $\oursol$ demonstrates significantly better performance in the unseen configurations, showing a performance gap ranging from 15.4\% to 70.8\%. 
In $\oursol$, the soft modularization combined with module selection facilitates effective module combinations for different tasks and constraints, rendering robust performance in unseen configurations.

\subsection{Ablation Study}

\begin{figure}[htbp] 
    \centering
    \begin{subfigure}{0.48\linewidth}
        \centering
        \includegraphics[width=\linewidth]{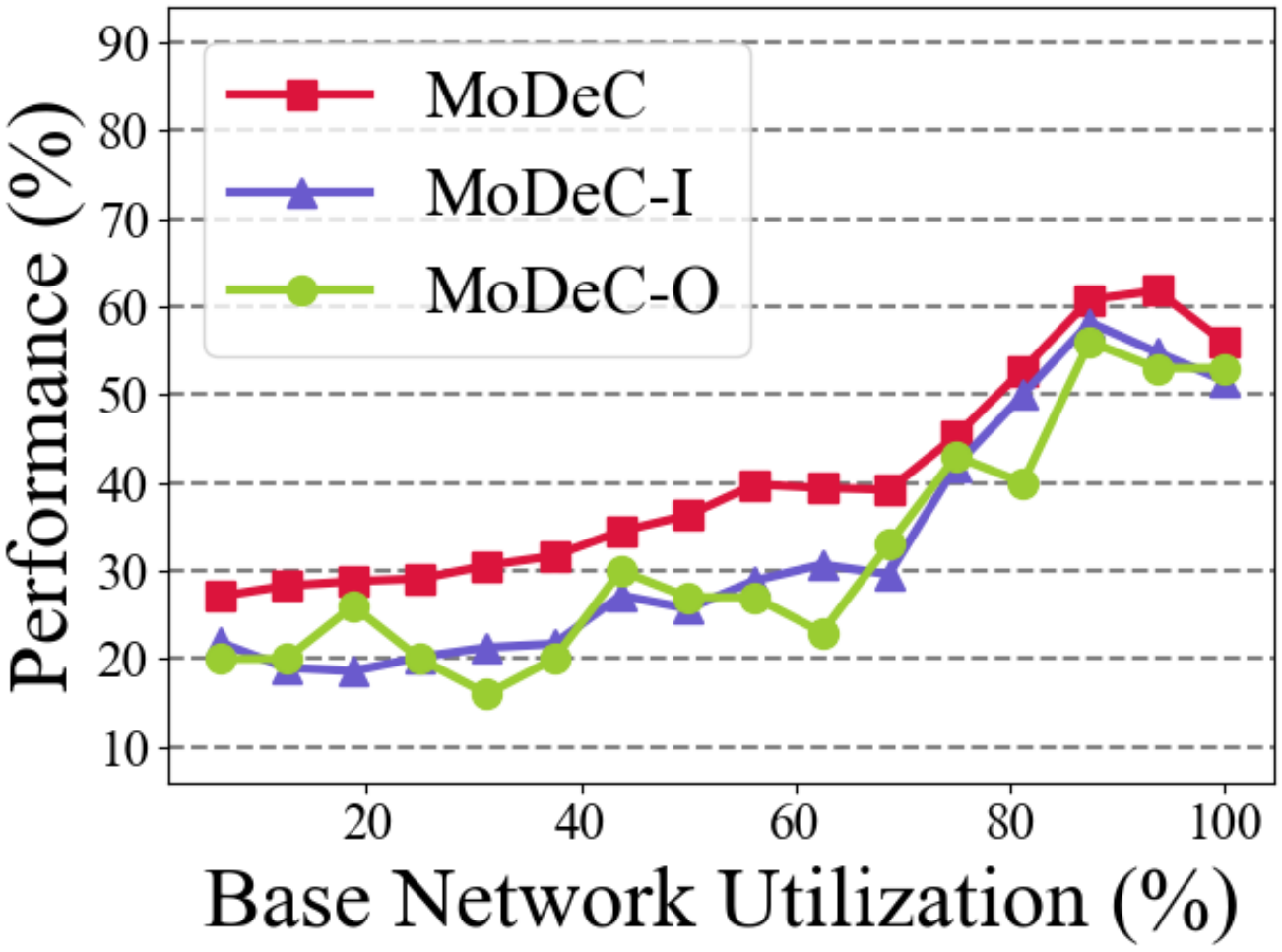} 
        \caption{Success Rate}
    \end{subfigure}
    \begin{subfigure}{0.48\linewidth}
        \centering
        \includegraphics[width=\linewidth]{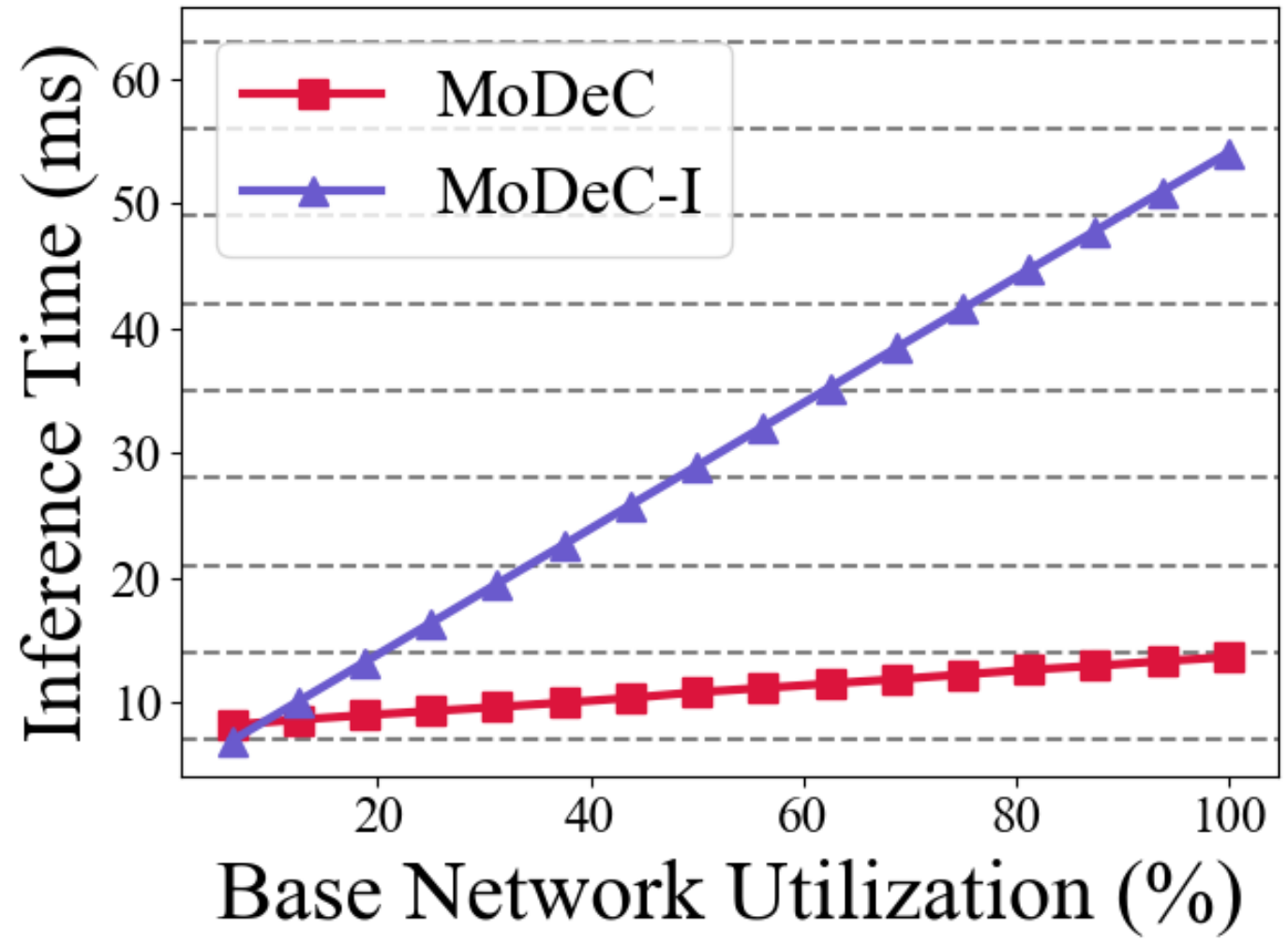} 
        \caption{Inference Time}
    \end{subfigure}
    \caption{Effect of Distillation}
    \label{fig:ablation_1}
\end{figure}

\noindent \textbf{Distillation.}
Figure~\ref{fig:ablation_1} clarifies the effects of distillation, where $\oursol$-I denotes a $\oursol$ variant without distillation, which adopts only the iterative module selection network; $\oursol$-O denotes another variant, which directly learns the single-step model selection (without distillation).
As shown, there is a significant difference in inference time between $\oursol$ and $\oursol$-I as the module utilization increases in (b), while $\oursol$ (with distillation) achieves higher performance in (a).
This is because the iterative module selection network is learned to infer as closely as possible to the original action under a limited module utilization, excluding environment rewards. 
When using both environment rewards and action distance, we observe a decline in the performance of the iterative module selective network. Due to changes in the environment reward, the actions of the iterative module selection network are not properly evaluated. 
The performance decline in $\oursol$-O stems from learning the module selection, which requires extensive exploration, yet is difficult in a single step without distillation.

\begin{figure}[htbp] 
        \centering
        \includegraphics[width=0.66\linewidth]{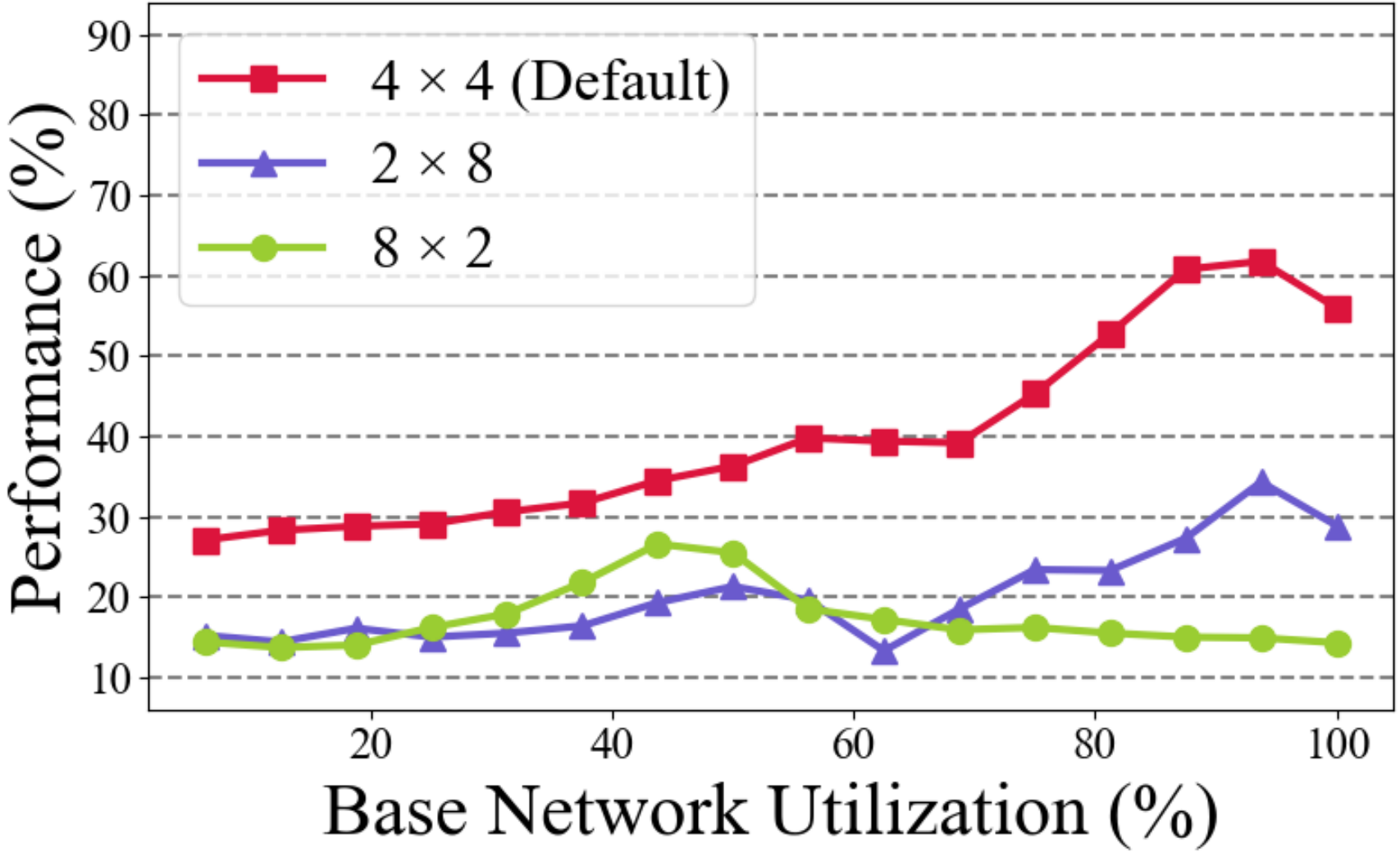} 
    \caption{Effect of Base Network Architecture}
    \label{fig:ablation_2}
    \vspace{-0.08in}
\end{figure}

\noindent \textbf{Base network architecture.}
Figure~\ref{fig:ablation_2} illustrates the effects of the modular base network architecture, where its number of layers can be configured differently. 
In our implementation, the base network has 4 layers with 4 modules per layer, each represented as $4 \times 4$ in the figure. We compare this with other variants, $2 \times 8$ and $8 \times 2$, in Meta-World. 
$\oursol$ shows the best performance by $4 \times 4$, which is a hyperparameter in the base network architecture.

\section{Conclusions}
In this work, we presented $\oursol$, which allows embodied agents to effectively adapt to time constraints on different target devices. The modular multi-task learning in $\oursol$ enables adaptive inference to a wide range of operational conditions including device resources, time constraints, and task specifications of embodied agents' multi-task settings, by dynamically adjusting the inference within a single model to satisfy the operational requirement. 
Through experiments with manipulation, autonomous driving, and object navigation scenarios of embodied agents, we verified that $\oursol$ is capable of handling those control tasks through rapid model adaptation to various operational conditions that can change over time. 
Our future work is to tackle the challenge of learning complex constraints from instructions, including safety and resource limitations.

\section*{Acknowledgements}
We would like to thank anonymous reviewers for their valuable comments and suggestions. 
This work was supported by 
Institute of Information \& communications Technology Planning \& Evaluation (IITP) grant funded by the Korea government (MSIT) (No.
2022-0-01045, 
2022-0-00043,
2021-0-00875,
2020-0-01821,
2019-0-00421) 
 and by the National Research Foundation of Korea (NRF) grant funded by the MSIT 
(No. RS-2023-00213118). 

\small
\bibliographystyle{ieeenat_fullname}
\bibliography{ref}

\end{document}